%% file: main.tex
\documentclass[10pt,twocolumn,letterpaper]{article}

\usepackage[pagenumbers]{wacv} %

\input{preamble}
\input{usercommand}
\newcommand{\methodname}{\textit{MagicDrive3D}\xspace}
\newcommand{\urlpath}{\url{https://flymin.github.io/magicdrive3d/}}

\definecolor{wacvblue}{rgb}{0.21,0.49,0.74}
\usepackage[pagebackref,breaklinks,colorlinks,allcolors=wacvblue]{hyperref}

\def\wacvPaperID{773} %
\def\confName{WACV}
\def\confYear{2026}

\title{\methodname: Controllable 3D Generation for Any-View Rendering \\ in Street Scenes}

\author{%
  Ruiyuan Gao$^{1}$,
  Kai Chen$^{2}$,
  Zhihao Li$^{3}$,
  Lanqing Hong$^{3\dag}$,
  Zhenguo Li$^{3}$,
  Qiang Xu$^{1\dag}$\\
$^{1}$CUHK
\enspace
$^{2}$HKUST
\enspace
$^{3}$Huawei Noah's Ark Lab\\
{\tt\small\{rygao,qxu\}@cse.cuhk.edu.hk},\enspace
{\tt\small kai.chen@connect.ust.hk}, \\
{\tt\small\{zhihao.li,honglanqing,li.zhenguo\}@huawei.com}
}

\begin{document}

\twocolumn[{%
\renewcommand\twocolumn[1][]{#1}%
\maketitle
\centering
    \vspace{-0.8cm}
    \input{section/teaser_content}
\vspace{0.5em}
\label{fig:teaser}
}]

\blfootnote{
$^{\dag}$Corresponding authors.
\\Project Page: \urlpath.}

\begin{abstract}
\vspace{-5mm}
\input{section/0.abs}
\vspace{-5mm}
\end{abstract}

\input{section/1.intro}
\input{section/2.related}

\input{section/3.methods}
\input{section/4.results}

\input{section/5.conclusion}

{
    \small
    \bibliographystyle{ieeenat_fullname}
    \bibliography{main}
}

\appendix
\input{section/appendix}

\end{document}

%% file: usercommand.tex
\usepackage{multirow}
\usepackage{pifont}
\usepackage{subcaption}
\usepackage{enumitem}
\usepackage{xspace}
\newcommand{\mb}[1]{\mathbf{#1}}

\usepackage{algorithm}
\usepackage{algorithmic}

\usepackage{array}
\newcolumntype{H}{>{\setbox0=\hbox\bgroup}c<{\egroup}@{}}

\def\maketitlesupplementary
   {
   \newpage
       {
        \centering
        \Large
        \vskip 0.5em
        Appendix (Supplementary Material) \\
        \vskip 1.0em
       }
   }

\makeatletter
\def\hlinew#1{%
\noalign{\ifnum0=`}\fi\hrule \@height #1 \futurelet
\reserved@a\@xhline}
\def\blfootnote{\xdef\@thefnmark{}\@footnotetext}
\makeatother

\makeatletter
\def\blfootnote{\xdef\@thefnmark{}\@footnotetext}
\DeclareRobustCommand\onedot{\futurelet\@let@token\@onedot}
\def\@onedot{\ifx\@let@token.\else.\null\fi\xspace}
\def\eg{\textit{e.g}\onedot}

\def\ie{\textit{i.e}\onedot}

\makeatother

\def\tabref#1{Table~\ref{#1}}
\def\figref#1{Figure~\ref{#1}}
\def\secref#1{Section~\ref{#1}}
\def\eqref#1{Equation~\ref{#1}}
\def\algref#1{Algorithm~\ref{#1}}
\def\appref#1{Appendix~\ref{#1}}

\definecolor{red}{rgb}{0.8,0,0}  
\definecolor{green}{RGB}{0, 133, 21}  
\definecolor{grey}{rgb}{0.5,0.5,0.5}  

%% file: section/teaser_content.tex
\includegraphics[width=0.98\textwidth]{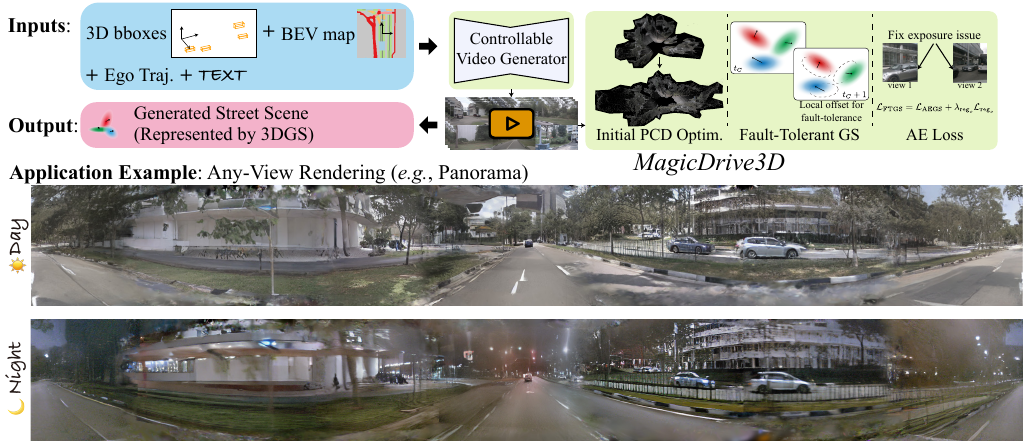}
    \vspace{-0.3cm}
    \captionof{figure}{Rendered panorama of the street scene generated from \methodname. With conditional controls from 3D bounding boxes of objects, BEV road map, ego trajectory, and text descriptions (\eg, timeofday), \methodname generates complex open-world 3D scenes represented by deformable Gaussians. }

%% file: section/0.abs.tex
Controllable generative models for images and videos have seen significant success, yet 3D scene generation, especially in unbounded scenarios like autonomous driving, remains underdeveloped. Existing methods lack flexible controllability and often rely on dense view data collection in controlled environments, limiting their generalizability across common datasets (\eg, nuScenes).
In this paper, we introduce \methodname, a novel framework for controllable 3D street scene generation that combines video-based view synthesis with 3D representation (3DGS) generation. It supports multi-condition control, including road maps, 3D objects, and text descriptions. Unlike previous approaches that require 3D representation before training, \methodname first trains a multi-view video generation model to synthesize diverse street views. This method utilizes routinely collected autonomous driving data, reducing data acquisition challenges and enriching 3D scene generation.
In the 3DGS generation step, we introduce Fault-Tolerant Gaussian Splatting to address minor errors and use monocular depth for better initialization, alongside appearance modeling to manage exposure discrepancies across viewpoints. Experiments show that \methodname generates diverse, high-quality 3D driving scenes, supports any-view rendering, and enhances downstream tasks like BEV segmentation, demonstrating its potential for autonomous driving simulation and beyond.

%% file: section/1.intro.tex
\section{Introduction}\label{sec:intro}

The advancement of generative models has enabled their use as data engines, particularly in the realm of controllable generative models, which can produce synthetic data without the need for relabeling~\cite{chen2024geodiffusion}. This capability is especially valuable in downstream applications like autonomous driving. However, much of the current research is limited to the generation of 2D data, such as images~\cite{chen2024geodiffusion,wang2024detdiffusion} and videos~\cite{gao2023magicdrive,gao2024magicdrivedit}, which falls short of meeting the demands for data that requires 3D information, such as viewpoint changes~\cite{tzofi2023view} or post-generation 3D editing~\cite{chen2025omnire}. Consequently, controllable generation of 3D street scenes remains an open challenge.

\input{tables/methods}

3D generation\footnote{In this paper, we focus on generative models where views/scenes are generated from latent variables, which is different from reconstruction (\eg, \cite{ziyang2023snerf}) or view synthesis (\eg, \cite{rombach2021geometry}).}, on the other hand, can directly produce 3D assets based on given conditions~\cite{tang2024dreamgaussian,yang2024hi3d}, supporting data engine applications that require 3D information.
However, many of them focus on single objects~\cite{tang2024dreamgaussian} or creative scenes~\cite{roblox2025cube} for gaming or content creation. Only a few attempts have been made at realistic scene generation for the physical world~\citep{kim2023neuralfield,bautista2022gaudi}; for example, NF-LDM~\citep{kim2023neuralfield} enables controllable synthesis of 3D scenes for autonomous driving.
The typical workflow for 3D generation involves first obtaining a compact 3D representation (\eg, voxel grids in \citep{kim2023neuralfield}) and then training a generative model based on this representation. Due to the high demands for data acquisition and reconstruction quality (\eg, the need for static scenes and numerous overlapping camera views), these methods are often limited to specific autonomous driving scenarios (\eg, parking lots, where no objects are moving), and thus do not support common street view datasets (\eg, nuScenes~\citep{nuscenes}).

These limitations prompt us to reconsider the relationship between the typical 3D generation workflow and 2D controllable generative models in autonomous driving scene generation. We observe that: \textbf{1)} 2D controllable generation requires less on data acquisition, as common roadside data can support training; \textbf{2)} 2D controllable generation provides the ability to manipulate data (e.g., generating multiple viewpoints of static scenes) to meet the data requirements of 3D reconstruction algorithms; \textbf{3)} 3D representation is the core of 3D generation to achieve consistent view rendering and novel view generalization.
These observations inspire us to restructure the 3D scene generation workflow by using 2D controllable generation as the initial step, leveraging its data manipulation capabilities to create reconstruction-friendly camera view data, and finally obtain the 3D representation of the scene. This ensures consistent viewpoint rendering and supports applications requiring 3D information (\eg, viewpoint augmentation~\cite{tzofi2023view}).

In this paper, we propose \methodname, a novel framework that combines video-based view synthesis and 3D representation (3DGS) generation for controllable 3D street scene generation, as illustrated in \figref{fig:teaser}.
Our approach begins with training a multi-view video generation model to synthesize multiple street views.
Besides text prompts, this model is configured using conditions from object boxes, road maps, and camera poses to support precise content control for static scene transformation.
Additionally, we incorporate ego trajectory embeddings representing the relative pose transformation between each frame, eliminating the need for pose estimation after view generation, leading to reconstruction-friendly view generation.
Next, we identify and solve several drawbacks in existing reconstruction algorithms on generated views. 
Since artifacts in view generation break the consistency assumption of reconstruction algorithms (\eg, 3DGS~\cite{kerbl3Dgaussians}), these artifacts can be amplified in the generated 3D representation, resulting in degraded rendering quality.
To solve this issue, we improve the 3D representation generation from the generated views from the perspectives of prior knowledge, modeling, and loss functions.
Most importantly, we introduce Fault-Tolerant Gaussian Splatting (FTGS) and appearance embedding maps to handle local inconsistency and exposure discrepancies, respectively.

Demonstrated by extensive experiments, our \methodname framework excels in generating highly realistic street scenes that align with road maps, 3D bounding boxes, and text descriptions, as exemplified in \figref{fig:teaser}.
We further show that the generated camera views can augment training of Bird's Eye View (BEV) segmentation tasks, improving the viewpoint robustness on perception tasks.
Notably, \methodname is the first to achieve controllable 3D street scene generation using a common driving dataset (e.g., the nuScenes dataset~\citep{nuscenes}), without requiring repeated data collection from static scenes or other sensor information.

We summarize our contributions as follows:
\begin{itemize}[leftmargin=*]
\item We propose \methodname, the first framework to effectively integrate both view synthesis and 3D representation generation for controllable 3D street scene generation. \methodname generates realistic 3DGS for street scenes that support rendering from any camera view according to various control signals.
\item We introduce a relative pose embedding to generate videos with accurate pose controls. Besides, we enhance the 3D representation generation quality through a dedicated pipeline (including Fault-Tolerant Gaussian Splatting) to handle local inconsistency and exposure discrepancies in the generated videos.
\item Through extensive experiments, we demonstrate that \methodname generates high-quality street scenes with various controllability. We also show that synthetic data improves 3D perception tasks, highlighting the practical benefits of our approach.
\end{itemize}

%% file: tables/methods.tex
\begin{table*}[t]
\centering
\caption{Comparisons on data collection requirements and functionality among different types of methods. \methodname belongs to 3D scene generation but can do more and requires less than previous works. Due to space limits, we only cite some typical works for each category. More detailed related works can be found in \secref{sec:related}.}
\label{tab:methods}
\vspace{-3mm}
\resizebox{\textwidth}{!}{
\begin{tabular}{l|cc|c|c|cc}
\toprule
\multirow{3}{*}{Method Type} &
\multicolumn{2}{c|}{Data Collection Requirements} &
\multirow{3}{*}{\begin{tabular}[c]{@{}c@{}}Prior \\ Control\end{tabular}} &
\multirow{3}{*}{\begin{tabular}[c]{@{}c@{}}Novel \\ Appearance\end{tabular}} &
\multicolumn{2}{c}{Novel Pose Rendering} \\ \cline{2-3} \cline{6-7} 
 & \multicolumn{1}{c|}{\begin{tabular}[c]{@{}c@{}}Allow \\ Dynamic Scene\end{tabular}} & 
 \multicolumn{1}{c|}{\begin{tabular}[c]{@{}c@{}} Allow Sparse \\ Pose Coverage\end{tabular}} &  & &
 \multicolumn{1}{c|}{Within Trajectory} &
 360$^{\circ}$ \\
 \midrule
Reconstruction~\cite{liu2024novel,wang2024freevs,yan2024streetcrafter,wei2024omni,zhao2024drivedreamer4d,ni2024recondreamer,zhou2024drivinggaussian,chen2025omnire} & \multicolumn{1}{c|}{\ding{52}} & \ding{52} & \ding{56} & \ding{56} & \multicolumn{1}{c|}{\ding{52}} & \ding{56} \\
Image/Video Generation~\cite{gao2023magicdrive,gao2024magicdrivedit} & \multicolumn{1}{c|}{\ding{52}} & \ding{52} & \ding{52} & \ding{52} & \multicolumn{1}{c|}{\ding{56}} & \ding{56} \\
3D Scene Generation (\cite{kim2023neuralfield,bautista2022gaudi}) & \multicolumn{1}{c|}{\ding{56}} & \ding{56} & \ding{52} & \ding{52} & \multicolumn{1}{c|}{\ding{52}} & \ding{52} \\
\midrule
3D Scene Generation (\methodname) & \multicolumn{1}{c|}{\ding{52}} & \ding{52} & \ding{52} & \ding{52} & \multicolumn{1}{c|}{\ding{52}} & \ding{52} \\ \bottomrule
\end{tabular}
}
\vspace{-5mm}
\end{table*}

%% file: section/2.related.tex
\section{Related Work}\label{sec:related}

\textbf{3D Scene Generation}.
Numerous 3D-aware generative models can synthesize images with explicit camera pose control~\citep{zhao2024pseudogeneralized,rombach2021geometry} and potentially other scene properties~\citep{tang2024dreamgaussian}, but only a few scale for open-ended 3D scene generation.
GSN~\citep{devries2021unconstrained} and GAUDI~\citep{bautista2022gaudi}, representative of models generating indoor scenes, utilize NeRF~\citep{mildenhall2020nerf} with latent code input for ``floorplan'' or tri-plane feature.
Their reliance on datasets covering different camera poses is incompatible with typical driving datasets where only fixed cameras are available.
NF-LDM~\citep{kim2023neuralfield} develops a hierarchical latent diffusion model for scene feature voxel grid generation.
However, their representation and complex modeling hinder fine detail generation.

Contrary to previous works focusing on scene generation using explicit geometry, often requiring substantial data not suitable for typical street view datasets (\eg, nuScenes~\citep{nuscenes}) as discussed in \secref{sec:intro}, we propose merging geometry-free view synthesis with geometry-focused scene representations for controllable street scene creation.
\tabref{tab:methods} shows a detailed comparison among different types of methods.
Methodologically, inpainting-based methods~\citep{chung2023luciddreamer,liang2024wonderland,yu2024wonderjourney} are most similar to our approach.
However, they generally rely on text-controlled image generation models, which cannot qualify as a view synthesis model.
Besides, they cannot complete our street scene generation task due to the lack of controllability, as showcased in \appref{app:baseline}.
In contrast, our video generation model is 3D-aware.
Besides, we propose several improvements over 3DGS for better scene generation quality.

\noindent\textbf{2D Street View Generation}.
Diffusion models~\citep{song2020score,ho2020denoising} boost a series of works for 2D street view generation.
These works span from single view video (\eg, \citep{wang2023drive}) to multi-view video (\eg, \citep{gao2023magicdrive,wen2023panacea,wang2023driving}).
Despite cross-view consistency being essential for multi-view video generation, their generalization ability on camera poses is limited due to the data-centric nature~\citep{gao2023magicdrive}. 
Furthermore, these models lack accurate control over frame transformation (\eg, precise ego trajectory), which is crucial for 3DGS generation.
Our work addresses this by enhancing pose control in video generation and proposing a dedicated 3DGS generation pipeline for geometric assurance.

\noindent\textbf{Street Scene Reconstruction}.
Scene reconstruction (or novel view rendering) for street views is useful in applications like driving simulation, and augmented and virtual reality.
For street scenes, attributes like scene dynamic and discrepancies from multi-camera data collection make typical large-scale reconstruction methods ineffective (\eg, \citet{rematas2022urf,martinbrualla2020nerfw,lin2024vastgaussian}). 
Hence, real data-based reconstruction methods like \citet{ziyang2023snerf,yan2024street} utilize LiDAR for depth prior, but their output only permits novel view rendering, lacking the capability of novel scene generation.
Unlike these methods, our approach synthesizes novel scenes under multiple levels of conditional controls.

%% file: section/3.methods.tex
\section{Preliminaries}\label{sec:preliminary}

\textbf{Problem Formulation}.
In this paper, we focus on controllable street scene generation.
Given scene description $\mb{S}_{t}$, our task is to generate street scenes (represented with 3D Gaussians $\mb{G}$) that correspond to the description from a set of latent $\mb{z}\sim\mathcal{N}(\mb{0},\mb{I})$, \ie $\mb{G}=\mathcal{G}(\mb{S}_{t},\mb{z})$.
To describe a street scene, we adopt the most commonly used conditions as per \citet{gao2023magicdrive,wang2023driving,wen2023panacea}.
Specifically, a frame of driving scene $\mb{S}_{t}=\{\mb{M}_{t}, \mb{B}_{t}, \mb{L}_{t}\}$ is described by road map $\mb{M}_{t}\in\{0, 1\}^{w\times h\times c}$ (a binary map representing a $w\times h$ meter road area in BEV with $c$ semantic classes), 3D bounding boxes $\mb{B}_{t}=\{(c_i, b_i)\}_{i=1}^{N}$ (each object is described by box $b_i=\{(x_j, y_j, z_j)\}_{j=1}^{8}\in\mathbb{R}^{8\times 3}$ and class $c_i\in\mathcal{C}$), and text $\mb{L}_{t}$ describing additional information about the scene (\eg, weather and time of day).
In this paper, we parameterize all geometric information according to the LiDAR coordinate of the ego car.

One direct application of scene generation is any-view rendering. Specifically, given any camera pose $\mb{P} = [\mb{K}, \mb{R}, \mb{t}]$ (\ie, intrinsics, rotation, and translation), the 3DGS model $\mb{G}(\cdot)$ should render photo-realistic views of the generated scene with 3D consistency, $\mathcal{I}^{r}=\mb{G}(\mb{P})$, which is not applicable to previous street view generation (\eg, \citet{gao2023magicdrive,wang2023driving,wen2023panacea}).
Besides, we present more practical applications in \secref{sec:exp}.

\begin{figure*}[t]
    \centering
    \includegraphics[width=0.95\linewidth]{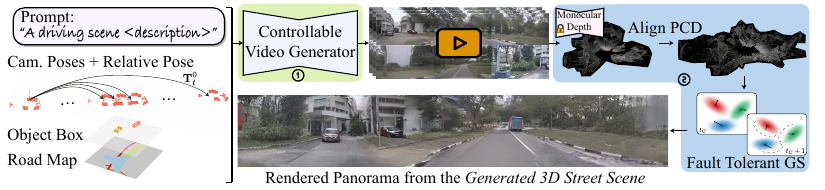}
    \vspace{-0.35cm}
    \caption{Method Overview of \methodname.
    For controllable street 3D scene generation, \methodname decomposes the task into two steps:
    \ding{172} conditional multi-view video generation, which tackles the control signals and generates consistent view priors to the novel scene;
    and \ding{173} Gaussian Splatting generation with our Enhanced GS pipeline, which supports various viewpoint rendering (\eg, panorama).
    }
    \label{fig:overview}
    \vspace{-0.5cm}
\end{figure*}

\noindent\textbf{3D Gaussian Splatting}.
We briefly introduce 3DGS since our scene representation is based on it.
3DGS~\citep{kerbl3Dgaussians} represents the geometry and appearance via a set of 3D Gaussians $\mb{G}$.
Each 3D Gaussian is characterized by its position $\pmb{\mu}_{p}$, anisotropic covariance $\pmb{\Sigma}_{p}$, opacity $\alpha_{p}$, and spherical harmonic coefficients for view-dependent colors $c_{p}$.
Given a sparse point cloud $\mathcal{P}$ and several camera views $\{\mathcal{I}_{i}\}$ with poses $\{\mb{P}_{i}\}$, a point-based volume rendering~\citep{zwicker2001ewa} is applied to make Gaussians optimizable through gradient descend and interleaved point densification.
Specifically, the loss is as \eqref{eq:lgs}, where $\mathcal{I}^{r}$ is the rendered image, $\lambda$ is a hyper-parameter, and $\mathcal{L}_{\text{D-SSIM}}$ denotes the D-SSIM loss~\citep{kerbl3Dgaussians}.
\begin{equation}
\mathcal{L}_{\text{GS}} = (1 - \lambda)\mathcal{L}_{1}(\mathcal{I}^{r}_{i}, \mathcal{I}_{i}) + \lambda \mathcal{L}_{\text{D-SSIM}}(\mathcal{I}^{r}_{i}, \mathcal{I}_{i})\label{eq:lgs}
\end{equation}

\section{Methods}\label{sec:method}

In this section, we introduce our controllable street scene generation pipeline.
Due to the challenges that exist in data collection, we integrate geometry-free view synthesis and geometry-focused 3DGS generation, detailed in \secref{sec:3D scene generation} and \figref{fig:overview}.
Specifically, we introduce a controllable video generation model to connect control signals and 3DGS generation by camera views generation (\secref{sec:video}).
Besides, we present a dedicated 3DGS generation pipeline for generated views, enhancing from prior, modeling, and loss perspectives (\secref{sec:gs}).

\subsection{Overview of \methodname}\label{sec:3D scene generation}

Direct modeling of controllable street scene generation faces two major challenges: scene dynamics and discrepancy in data collection.
\textit{Scene dynamics} refers to the movements and deformation of elements in the scene, while \textit{discrepancy in data collection} refers to the discrepancy (\eg, exposure) caused by data collection.
These two challenges are even more severe due to the sparsity of cameras for street views (\eg, typically only 6 surrounding perspective cameras).
Therefore, reconstruction-generation frameworks do not work well for street scene generation~\citep{kim2023neuralfield,bautista2022gaudi}.

\figref{fig:overview} shows the overview of \methodname.
Given scene descriptions $\mb{S}$ as input, \methodname first extend the descriptions into sequence $\{\mb{S}_{t}\}$, where $t\in[0,T]$ according to preset camera poses $\{\mb{P}_{c,t}\}$, and generate a sequence of successive multi-view images $\{\mathcal{I}_{c,t}\}$, where $c\in\{1,\dots,N\}$ refers to $N$ surrounding cameras, according to conditions $\{\mb{S}_{t},\mb{P}_{c,t}\}$ (detailed in \secref{sec:video}).
Then we generate 3D Gaussian representation of the scene with $\{\mathcal{I}_{c,t}\}$ and camera poses $\{\mb{P}_{c,t}\}$ as input.
This step contains an initializing procedure with a pre-trained monocular depth model and an optimizing process with Fault-Tolerant Gaussian Splatting (FTGS, detailed in \secref{sec:gs}).
Consequently, the generated street scene reflects different control signals accurately and supports any-view rendering.

\methodname integrates geometry-free view synthesis and GS-based scene representation, which are geometry-focused.
Specifically, control signals are tackled by a multi-view video generator, while the 3DGS generation step guarantees the generalization ability for any-view rendering.
Such a video generator has two advantages:
first, since multi-view video generation does not require generalization on novel views~\citep{gao2023magicdrive}, it poses less data dependency for street scenes;
second, through conditional training, the model is able to decompose control signals, turning dynamic scenes into static scenes that are more friendly for 3DGS generation.
Besides, for the 3DGS generation step, strong priors from the multi-view video reduce the burden for scene modeling with complex details.

\begin{figure*}[t]
    \centering
    \includegraphics[width=0.7\linewidth]{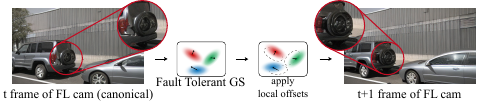}
    \vspace{-0.35cm}
    \caption{Illustration of the local inconsistency from two successive generated frames of Front-Left (FL) camera. Even though our video generation model retains fine 3D consistency, minor discrepancies are inevitable. Our FTGS can effectively reconstruct the scene with awareness of such discrepancy.}
    \label{fig:local_dyn}
    \vspace{-0.65cm}
\end{figure*}

\subsection{Video Generation with Relative Pose Control}\label{sec:video}

Given scene descriptions and a sequence of camera poses $\{\mb{S}_{t},\mb{P}_{c,t}\}$, our video generator is responsible for multi-view video generation.
Although many previous arts for street view generation achieve expressive visual effects (e.g., ~\citep{gao2023magicdrive,wen2023panacea,wang2023driving,wang2023drive}), their formulations leave out a crucial requirement for 3D modeling.
Specifically, the camera pose $\mb{P}_{c,t}$ they use is typically relative to the LiDAR coordinate of each frame.
Thus, there is no precise control signal related to the ego trajectory, which significantly determines the geometric relationship between views of different $t$s.

In our video generation model, we amend such precise control ability by adding the transformation between each frame to the first frame, \ie, $\mb{T}^{0}_{t}$.
To properly encode such information, we adopt Fourier embedding with Multi-Layer Perception (MLP), and concatenate the embedding with the original embedding of $\mb{P}_{c,t}$, similar to \citet{gao2023magicdrive}.
As a result, our video generator provides better temporal quality across frames, most importantly, making the camera poses to each view available in the same coordinate, \ie, $[\mb{R}^0_{c,t},\mb{t}^{0}_{c,t}]=\mb{T}^{0}_{t}[\mb{R}_{c,t},\mb{t}_{c,t}]$.

\begin{figure}[t]
    \centering
    \includegraphics[width=0.85\linewidth]{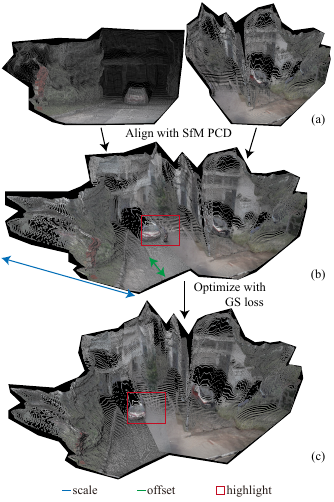}
    \vspace{-0.4cm}
    \caption{
    We optimize the monocular depths (a) with 2 steps for better alignment: coarse scale/offset estimation with SfM PCD (b) and GS optimization (c).
    }
    \label{fig:depth opt}
    \vspace{-0.55cm}
\end{figure}

\subsection{Enhanced FTGS for 3DGS Generation}\label{sec:gs}
As introduced in \secref{sec:preliminary}, 3DGS is a flexible explicit representation for 3D scenes.
Besides, the fast training and rendering speed of 3DGS makes it highly suitable for reducing generation costs in our scene creation pipeline.
However, similar to other 3D reconstruction methods, 3DGS necessitates high cross-view 3D consistency at the pixel level, which unavoidably magnifies the minor artifacts in the generated data into conspicuous errors. 
Therefore, we propose improvements for 3DGS from the perspectives of \textit{prior}, \textit{modeling}, and \textit{loss}, enabling 3DGS to tolerate minor artifacts in the generated camera view from our video generation model, thereby becoming a potent tool for enhancing geometric consistency in rendering.

\noindent\textbf{Prior: Consistent Depth Prior}.
As essential geometry information, depth is extensively utilized in street scene reconstruction, such as the depth value from LiDAR or other depth sensors used by \citet{yan2024street}.
However, for synthesized camera views, LiDAR/depth information is unavailable (also making these methods not suitable for our scenario).
Therefore, we propose to use a pre-trained monocular depth estimator~\citep{bhat2023zoedepth} to infer depth information.

While monocular depth estimation is separate for each camera view, proper scale $s_{c,t}$ and offset $b_{c,t}$ parameters should be estimated to align them for a single scene~\citep{zhou2024dynpoint}, as in \figref{fig:depth opt}(a).
To this end, we first apply the Point Cloud (PCD) from Structure of Motion (SfM)~\citep{schoenberger2016mvs,schoenberger2016sfm} for estimation, shown in \figref{fig:depth opt}(b).
However, such PCD is too sparse to accurately restore $(s_{c,t}, b_{c,t})$ for any views.
To bridge the gap, secondly, we propose further optimizing the $(s_{c,t}, b_{c,t})$ using the GS loss, as in \figref{fig:depth opt}(c).
Specifically, we replace the optimization for Guassian centers $\pmb{\mu}_{i}$ with $(s_{c,t}, b_{c,t})$.
After the optimization, we reset $\pmb{\mu}_{i}$ with points from depth values.
Since GS algorithm is sensitive to accurate point initialization~\citep{kerbl3Dgaussians,fan2024instantsplat}, our method provides a useful prior for 3DGS generation in this sparse view scenario.

\begin{figure*}[t]
    \centering
    \includegraphics[width=0.98\linewidth]{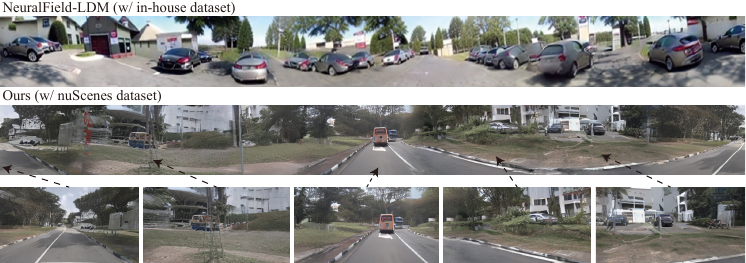}
    \vspace{-0.3cm}
    \caption{Qualitative comparison with NF-LDM~\citep{kim2023neuralfield}.
    Our method can generate higher quality 3D scenes while maintaining better object geometry, with stronger controllability compared to NF-LDM (see the significant object deformation in NF-LDM).
    Panoramas for GS are transformed and stitched from perspective cameras with $90^\circ$ FOV.
    Views in the last row are rendered with unseen camera rigs of nuScenes.}
    \label{fig:comparison}
    \vspace{-0.5cm}
\end{figure*}

\noindent\textbf{Modeling: Fault-Tolerant Gaussians for Local Inconsistency}.
Despite the 3D geometric consistency provided by our video generation model, there are inevitably pixel-level disparities in some object details, as shown in \figref{fig:local_dyn}.
Even worse, the strict consistency assumption of 3DGS~\citep{kerbl3Dgaussians} (and all other reconstruction algorithms, \eg, \citep{zhou2024drivinggaussian,ziyang2023snerf,Wu_2024_CVPR}) can amplify these minor errors, resulting in floater artifacts.
To mitigate the impact of these errors, we propose Fault-Tolerant Gaussian Splitting (FTGS), which reduces the requirement for temporal consistency between frames, thereby ensuring the 3DGS generation quality in the canonical space from the generated views.

Specifically, as shown in \figref{fig:local_dyn}, we pick the center frame $t=t_{C}$ as the canonical space and enforce all Gaussians in this space.
Then, we allocate a set of offsets to each Gaussian, $\pmb{\mu}^{o}_{p}(t)\in\mathbb{R}^{3}$, where $t\in [1,\dots,T]$ and $\pmb{\mu}^{o}_{p}(t_{C})\equiv\mb{0}$.
Note that, different camera views from the same $t$ share the same $\pmb{\mu}^{o}_{p}(t)$ for each Gaussian, and we apply regularization on them to keep the offsets in local, as shown in \eqref{equ:reg}:
\begin{equation}
    \mathcal{L}_{\text{reg}_{o}} = \Vert \pmb{\mu}^{o}(t)\Vert_{2}. \label{equ:reg}
\end{equation}
Consequently, $\pmb{\mu}^{o}_{p}(t)$ can manage the local inconsistency driven by pixel-level disparities, while $\pmb{\mu}$ focuses on the global geometric correlations.
It ensures the quality of scene 3DGS generation by leveraging consistent parts across different viewpoints, simultaneously eliminating artifacts.
Besides, with the analytical gradient w.r.t. $SE(3)$ pose of cameras~\citep{Matsuki:Murai:etal:CVPR2024}, we make the camera pose optimizable in the final few steps of GS iterations, which helps to mitigate the local inconsistency from camera poses.

\noindent\textbf{Loss: Aligning Exposure with Appearance Modeling}.
Typical street view datasets are collected using multiple cameras, each operating independently with auto-exposure and auto-white-balance~\citep{nuscenes}. Appearance modeling addresses issues from such in-the-wild reconstructions~\citep{martinbrualla2020nerfw}. Similarly, video generation is optimized to match the real data distribution, so differences between cameras also appear in generations. We propose a dedicated appearance modeling technique for GS representation as a solution.

We hypothesize that the disparity between different views can be represented by affine transformations $\mb{A}_{i}(\cdot)$ for $i$-th camera view.
An Appearance Embedding (AE) map $\mb{e}_{i}\in\mathbb{R}^{w_{e}\times h_{e}\times c_{e}}$ is allocated for each view, and a Convolutional Neural Network (CNN) is utilized to approximate this transformation matrix $w_{\mb{A}}\in\mathbb{R}^{w \times h\times 3}$ (\appref{app:arch} contains more details).
The final computation of the pixel-wise $\ell_{1}$ loss is conducted using the transformed image.

Therefore, our final loss for FTGS is as \eqref{equ:loss}, where $\lambda_{{\text{reg}_{o}}}$ is the hyper-parameter for offset regularization.
\begin{equation}
{\scriptsize
\begin{aligned}
\mathcal{L}_{\text{FTGS}} &= 
\mathcal{L}_{\text{AEGS}} + \lambda_{{\text{reg}_{o}}}\mathcal{L}_{{\text{reg}_{o}}} \\
&=(1 - \lambda)\mathcal{L}_{1}(\mb{A}_{i}(\mathcal{I}^{r}_{i}), \mathcal{I}_{i}) + \lambda \mathcal{L}_{\text{D-SSIM}}(\mathcal{I}^{r}_{i}, \mathcal{I}_{i}) + \lambda_{{\text{reg}_{o}}}\mathcal{L}_{{\text{reg}_{o}}}  
\end{aligned}
\label{equ:loss}
}\end{equation}
\appref{app:optimflow} holds a full description of the algorithm.

%% file: section/4.results.tex
\begin{table}
\centering
\setlength\tabcolsep{3pt}
\resizebox{\linewidth}{!}{
\begin{tabular}{l|ccc}
\toprule
Methods & FVD & FID (seen) & FID (novel) \\
\midrule
MagicDrive &  177.26 & 20.92  & N/A \\
\methodname (video gen.) & 164.72 & 20.67 & N/A \\
\midrule
3DGS & N/A &  45.07 & 145.72  \\
\methodname (scene gen.) & N/A & 23.99 & 34.45 \\
\bottomrule
\end{tabular}
}
\vspace{-0.33cm}
\caption{Generation quality evaluation. We adopt all val. scenes from nuScenes. All generated views are used for 3DGS generation.
Novel views using camera poses different from nuScenes.}
\label{tab:fid}
\vspace{-0.57cm}
\end{table}

\section{Experiments}\label{sec:exp}

\subsection{Experimental Setup}\label{sec:exp setup}
\textbf{Dataset}.
We test \methodname using nuScenes dataset~\citep{nuscenes}, which is commonly used for generating and reconstructing street views~\citep{gao2023magicdrive,wen2023panacea,wang2023driving,ziyang2023snerf}.
The official configuration is followed (700 street-view video clips of $\sim$20s each for training and another 150 clips for validation).
For semantics in control signals, we follow \citet{gao2023magicdrive}, using 10 object classes and 8 road classes.

\noindent\textbf{Metrics and Settings}.
There is no direct metric for 3D scene generation.
Following \citet{kim2023neuralfield}, we primarily evaluate \methodname using the Fréchet Inception Distance (FID) by rendering novel views unseen in the dataset and comparing their FID with real images.

To further demonstrate the effectiveness of each component within our framework, we conducted ablated studies on the improvements in video generation capabilities and the enhancements in content reconstruction quality.
The method's video generation ability is evaluated using Fréchet Video Distance (FVD).
\appref{app:more details} holds more details.

\begin{figure*}[t]
    \centering
    \includegraphics[width=0.9\linewidth]{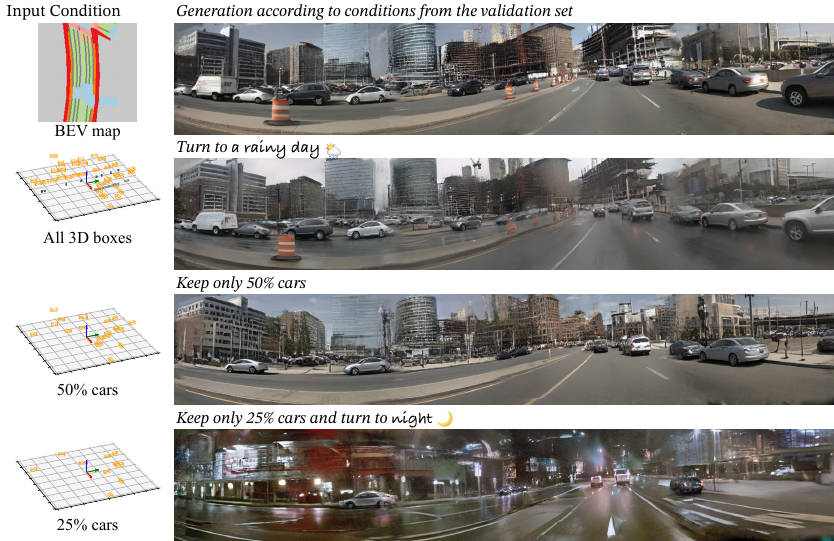}
    \vspace{-0.3cm}
    \caption{Qualitative evaluation for controllability (we show the view from back-left to front-right area).
    By changing different control signals, \methodname can edit the scene from different levels.
    }
    \vspace{-0.55cm}
    \label{fig:editing}
\end{figure*}

\subsection{Main Results}

\textbf{Generation Quality}.
In \tabref{tab:fid}, generation quality is evaluated in two ways.
First, 3D scene quality is assessed by FID of rendered views (last row of \tabref{tab:fid}), with qualitative comparisons in \figref{fig:comparison}. Lacking a quantitative baseline, we compare our pipeline's renderings with 3DGS. The last two rows of \tabref{tab:fid} show that our enhanced FTGS improves visual quality, especially in novel views.
Second, video quality is assessed using extended 16-frame MagicDrive~\citep{gao2023magicdrive} as a baseline. \methodname significantly boosts video quality (shown by FVD), proving the effectiveness of the proposed relative camera pose embedding for enhancing temporal quality.
More qualitative comparisons can be found in \appref{app:baseline}.

\begin{figure}[t]
     \centering
     \begin{subfigure}[b]{0.48\linewidth}
         \centering
         \includegraphics[height=2cm]{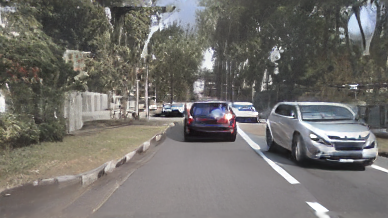}
         \vspace{-0.1cm}
         \caption{Original view.}
         \label{fig:mv_ori}
     \end{subfigure}
     \hfill
     \begin{subfigure}[b]{0.48\linewidth}
         \centering
         \includegraphics[height=2cm]{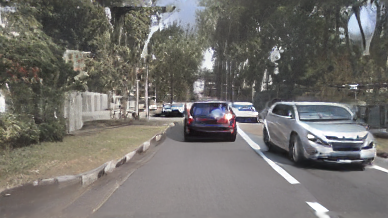}
         \vspace{-0.1cm}
         \caption{Move forward 0.1m.}
         \label{fig:mv_0}
     \end{subfigure}
     \hfill
     \begin{subfigure}[b]{0.48\linewidth}
         \centering
         \includegraphics[height=2cm]{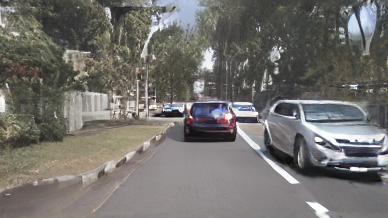}
         \vspace{-0.1cm}
         \caption{Move forward 1m.}
         \label{fig:mv_3}
     \end{subfigure}
     \hfill
     \begin{subfigure}[b]{0.48\linewidth}
         \centering
         \includegraphics[height=2cm]{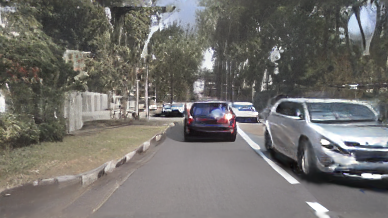}
         \vspace{-0.1cm}
         \caption{Move forward 1.6m.}
         \label{fig:mv_5}
     \end{subfigure}
     \vspace{-0.3cm}
     \caption{
     Application of rendering object-level dynamics. After scene generation, we can segment and move the vehicle (the one on the right) in 3D to render a dynamic object.}
    \label{fig:mv_object}
    \vspace{-0.55cm}
\end{figure}

\noindent\textbf{Controllability}.
\methodname accepts 3D bounding boxes, BEV map, and text as control signals, each of which possesses the capacity to independently manipulate the scene.
\figref{fig:editing} shows such controllability, where we edit a scene from the nuScenes validation set.
Clearly, \methodname effectively alters the scene generation to align with various control signals.

Furthermore, \secref{sec:train} shows that the rendered views of the generated scenes can be used as training data to improve the performance of perception models. Such results can serve as further evidence of the controllability.

\input{tables/cvt}

\subsection{Data Augmentation for Perception Tasks}\label{sec:train}
We demonstrate that street 3D scene generation can serve as a data engine for perception tasks, leveraging the advantage of any-view rendering ability of our method to improve viewpoint robustness for perception models~\citep{tzofi2023view}.
We employ CVT~\citep{zhou2022cross} and the BEV segmentation task following the evaluation protocols of \citet{zhou2022cross,gao2023magicdrive}.
By incorporating 4 different rigs on the FRONT camera and adding rendered views for training, the negative impact from viewpoint changes is alleviated (\tabref{tab:cvt}), exemplifying the utility of street scene generation in training perception tasks.

\input{tables/recon}

\subsection{Application: Render Object-level Dynamic}\label{sec:app-dyn}
\methodname is capable of generating 3DGS representations of scenes, which facilitates various applications in scene editing. Our method utilizes metric scale modeling, guaranteeing that any modifications made to the scenes accurately reflect real-world physical distances. As illustrated in \figref{fig:mv_object}, we segmented the generated GS and repositioned the object on the right.
The resulting scene GS supports rendering effectively.

\input{tables/ablation}

\subsection{Ablation Study}\label{sec:ablation}
\noindent\textbf{Effectiveness of the Enhanced GS pipeline}.
We first show the effectiveness of the proposed enhanced GS pipeline through an experimental setting similar to the evaluation of a general reconstruction method~\cite{kerbl3Dgaussians}, \ie, comparing renderings with ground truth (GT) images, with 3DGS~\cite{kerbl3Dgaussians} as the baseline method.
Specifically, two testing scenarios are employed (as shown in \tabref{tab:settings}, appendix)
: 360$^\circ$, where all six views from $t=9$ are reserved for testing; and vary-t, where one view is randomly sampled from different $t$ to assess long-range reconstruction ability through $t$.
Here, the views generated by the video generator are treated as GT.
Performance is assessed using L1, PSNR, SSIM~\citep{wang2004image}, and LPIPS~\citep{zhang2018perceptual}, as per \citet{kerbl3Dgaussians}.

Among all the metrics in \tabref{tab:recon}, our enhanced GS not only improves reconstruction quality for training views but also drastically enhances generalization quality for testing views, compared to 3DGS.
We also include more comparison with 4DGS~\citep{Wu_2024_CVPR} as another baseline in \appref{app:4dgs}.

\noindent\textbf{Ablation on Enhanced Gaussian Splatting}.
As detailed in \secref{sec:gs}, three enhancements—prior, modeling, and loss—were made to 3DGS. To evaluate their efficacy, each was ablated from the final algorithm, with results shown in \tabref{tab:ablation}. Notations ``w/o depth scale opt.'' and ``w/o depth opt.'' indicate the absence of GS loss optimization for \((s_{c,t}, b_{c,t})\) and the use of direct output from the monocular depth model, respectively. Removing each component reduced performance, and incorrect depth sometimes performed worse than the 3DGS baseline. Removing AE in ``vary-t'' led to lower PSNR but improved LPIPS, as AE mitigates pixel-wise color constraints during 3DGS generation.

\appref{app:abl} contains more ablation studies.

%% file: tables/cvt.tex
\begin{table*}[t]
\centering
\setlength\tabcolsep{3pt}
\vspace{-0.2cm}
\begin{tabular}{@{}c|l|Hccccc@{}} %
\toprule
Setting & Method & nuScene val & no rig & depth+0.5m & pitch-5$^\circ$ & yaw+5$^\circ$ & yaw-5$^\circ$ \\ \midrule
\multirow{3}{*}{vehicle} & only real data &
    32.91 & 17.14 & 16.63 & 15.50 & 16.99 & 15.94 \\ %
 & w/ render view (no rig) &
    32.93 {\small\textcolor{green}{+0.02}} &
    20.67 {\small\textcolor{green}{+3.53}} &
    20.13 {\small\textcolor{green}{+3.50}} &
    17.03 {\small\textcolor{green}{+1.53}} &
    19.40 {\small\textcolor{green}{+2.41}} &
    19.30 {\small\textcolor{green}{+3.36}}  \\ %
 & w/ random aug. of 4 rigs &
    \textbf{32.67 {\small\textcolor{red}{-0.24}}} &
    \textbf{21.05 {\small\textcolor{green}{+3.91}}} &
    \textbf{20.46 {\small\textcolor{green}{+3.83}}} &
    \textbf{19.75 {\small\textcolor{green}{+4.25}}} &
    \textbf{19.81 {\small\textcolor{green}{+2.82}}} &
    \textbf{19.83 {\small\textcolor{green}{+3.89}}} \\ %
 \midrule
\multirow{3}{*}{road} & only real data &
    73.69 & 54.94 & 54.56 & 53.82 & 54.20 & 53.67 \\ %
 & w/ render view (no rig) & 
    71.32 {\small\textcolor{red}{-2.36}} &
    60.31 {\small\textcolor{green}{+5.37}} &
    59.93 {\small\textcolor{green}{+5.37}} &
    58.46 {\small\textcolor{green}{+4.64}} &
    59.16 {\small\textcolor{green}{+4.96}} &
    59.32 {\small\textcolor{green}{+5.65}} \\ %
 & w/ random aug. of 4 rigs &
    \textbf{71.69 {\small\textcolor{red}{-2.00}}} &
    \textbf{60.59 {\small\textcolor{green}{+5.65}}} &
    \textbf{60.38 {\small\textcolor{green}{+5.82}}} &
    \textbf{59.95 {\small\textcolor{green}{+6.13}}} &
    \textbf{60.21 {\small\textcolor{green}{+6.01}}} &
    \textbf{60.29 {\small\textcolor{green}{+6.62}}} \\ %
\bottomrule
\end{tabular}
\vspace{-0.3cm}
\caption{\methodname improves the viewpoint robustness~\citep{tzofi2023view} of CVT~\citep{zhou2022cross}. All results are mIoU for BEV segmentation.
Colors highlight the differences with baseline.
The best results are in \textbf{bold}.
}
\label{tab:cvt}
\vspace{-0.5cm}
\end{table*}

%% file: tables/recon.tex
\begin{table}[tb]
\centering
\setlength\tabcolsep{1.8pt}
\scalebox{0.95}{
\begin{tabular}{@{}ccl|cccc@{}}
\toprule
\multicolumn{2}{c|}{Settings} & Methods & L1 $\downarrow$ & PSNR $\uparrow$ & SSIM $\uparrow$ & LPIPS $\downarrow$ \\ \midrule
\multicolumn{1}{c|}{\multirow{6}{*}{\begin{tabular}[c]{@{}c@{}}train\\view\end{tabular}}} &
\multicolumn{1}{c|}{\multirow{3}{*}{vary-t}} & 3DGS & 0.0189 & 30.1191 & 0.9261 & 0.1259 \\
\multicolumn{1}{c|}{} & \multicolumn{1}{c|}{} & 3DGS + cc & 0.0186 & 30.2498 & 0.9253 & 0.1258 \\
\multicolumn{1}{c|}{} & \multicolumn{1}{c|}{} & \textbf{Ours} & \textbf{0.0167} & \textbf{32.6001} & \textbf{0.9544} & \textbf{0.0673} \\
\cmidrule(l){2-7} 
\multicolumn{1}{c|}{} & \multicolumn{1}{c|}{\multirow{3}{*}{360$^{\circ}$}} & 3DGS & 0.0202 & 29.4943 & 0.9187 & 0.1365 \\
\multicolumn{1}{c|}{} & \multicolumn{1}{c|}{} & 3DGS + cc & 0.0199 & 29.6327 & 0.9178 & 0.1366 \\
\multicolumn{1}{c|}{} & \multicolumn{1}{c|}{} & \textbf{Ours} & \textbf{0.0174} & \textbf{32.2104} & \textbf{0.9530} & \textbf{0.0693} \\ \midrule
\multicolumn{1}{c|}{\multirow{6}{*}{\begin{tabular}[c]{@{}c@{}}test\\view\end{tabular}}} & \multicolumn{1}{c|}{\multirow{3}{*}{vary-t}} & 3DGS & 0.0890 & 17.9879 & 0.4378 & 0.4648 \\
\multicolumn{1}{c|}{} & \multicolumn{1}{c|}{} & 3DGS + cc & 0.0799 & 19.1387 & 0.4814 & 0.4697 \\
\multicolumn{1}{c|}{} & \multicolumn{1}{c|}{} & \textbf{Ours} & \textbf{0.0738} & \textbf{19.7063} & \textbf{0.5145} & \textbf{0.4115} \\
\cmidrule(l){2-7} 
\multicolumn{1}{c|}{} & \multicolumn{1}{c|}{\multirow{3}{*}{360$^{\circ}$}} & 3DGS & 0.0910 & 17.8322 & 0.4318 & 0.4756 \\
\multicolumn{1}{c|}{} & \multicolumn{1}{c|}{} & 3DGS + cc & 0.0804 & 19.0773 & 0.4777 & 0.4796 \\
\multicolumn{1}{c|}{} & \multicolumn{1}{c|}{} & \textbf{Ours} & \textbf{0.0622} & \textbf{21.0351} & \textbf{0.5754} & \textbf{0.3207} \\
\bottomrule
\end{tabular}
}
\vspace{-0.3cm}
\caption{Performance comparison between the enhanced GS and its baseline. We randomly sample 100 scenes from the nuScenes val. set for evaluation. ``cc'' refers to color correction from~\cite{barron2022mipnerf360}. Although 3DGS does not consider appearance differences, we apply ``cc'' to it for fair comparisons.}
\label{tab:recon}
\vspace{-0.4cm}
\end{table}

%% file: tables/ablation.tex
\begin{table}[t]
\centering
\setlength\tabcolsep{2.8pt}
\scalebox{0.95}{
\begin{tabular}{@{}l|cccc@{}}
\toprule
Methods & L1 $\downarrow$ & PSNR $\uparrow$ & SSIM  $\uparrow$ & LPIPS $\downarrow$ \\ \midrule
\multicolumn{5}{l}{vary-t setting:} \\\midrule
3DGS & 0.0799 & 19.1387 & 0.4814 & 0.4697 \\
w/o AE & 0.0822 & 18.8467 & 0.4758 & 0.4452 \\
w/o depth scale opt. & 0.0815 & 18.8885 & 0.4767 & 0.4366 \\
w/o depth opt. & 0.1046 & 17.2776 & 0.4399 & 0.5545 \\
w/o xyz offset + cam & 0.0798 & 19.1657 & 0.4919 & 0.4580 \\
\textbf{Ours} & \textbf{0.0738} & \textbf{19.7063} & \textbf{0.5145} & \textbf{0.4115} \\ \midrule

\multicolumn{5}{l}{360$^\circ$ setting:} \\\midrule
3DGS & 0.0804 & 19.0773 & 0.4777 & 0.4796 \\
w/o AE & 0.0722 & 19.7742 & 0.5114 & 0.3791 \\
w/o depth scale opt. & 0.0736 & 19.6501 & 0.5086 & 0.3736 \\
w/o depth opt. & 0.0995 & 17.6707 & 0.4487 & 0.5150 \\
w/o xyz offset + cam & 0.0798 & 19.1682 & 0.4888 & 0.4663 \\
\textbf{Ours} & \textbf{0.0622} & \textbf{21.0351} & \textbf{0.5754} & \textbf{0.3207} \\ \bottomrule
\end{tabular}
}
\vspace{-0.3cm}
\caption{Ablation study on each component of the enhanced GS, with settings in \tabref{tab:settings} (appendix) and 100 random scenes from the nuScenes validation set.}
\label{tab:ablation}
\vspace{-0.65cm}
\end{table}

%% file: section/5.conclusion.tex
\section{Conclusion and Discussion}\label{sec:conclusion}

This paper introduces \methodname, a novel framework for controllable 3D street scene generation that integrates multi-view video synthesis with 3D representations. \methodname uses a relative pose embedding in video generation to enhance inter-frame quality while providing camera pose data for each view. The enhanced fault-tolerant GS in \methodname improves 3DGS generation quality from the generated camera views. Consequently, \methodname significantly reduces data collection requirements, enabling training on standard autonomous driving datasets such as nuScenes. Comprehensive experiments demonstrate that \methodname can produce high-quality 3D street scenes with multi-level controls. Furthermore, the generated scenes enhance viewpoint robustness in perception tasks like BEV segmentation.

%% file: section/appendix.tex
\clearpage
\onecolumn
\setcounter{page}{1}
\setcounter{section}{0}
\renewcommand{\thesection}{\Alph{section}}
\setcounter{figure}{0}
\renewcommand{\thefigure}{\Roman{figure}}
\setcounter{table}{0}
\renewcommand{\thetable}{\Roman{table}}
\maketitlesupplementary

\begin{center}
\noindent Please find the video visualizations on our project website: \urlpath   
\end{center}

\section{Further Discussion on Differences from Reconstruction Works}
The main focus of this work is the generation of non-existent 3D scenes, which distinguishes it from the reconstruction of existing scenes (such as~\cite{yan2024street, zhou2024drivinggaussian}). While 3D scene reconstruction also leverages video generation models to enhance quality~\cite{ni2024recondreamer, yan2024streetcrafter}, its goal is to produce novel views of the same scene from a given set of camera angles. In contrast, our model can directly generate entirely new 3D scenes through latent variable sampling after training.
The key difference is that our approach enables the creation of entirely new scenes, a capability that reconstruction methods cannot achieve (as shown in \tabref{tab:methods}). Specifically, our model can generate novel scenes that, while adhering to control conditions, allow for the updating of objects and environmental features, such as buildings, across different samples. Figures~\ref{fig:teaser} and~\ref{fig:editing} demonstrate our method's ability to construct diverse 3D scenes based on varying conditions.
We hope this discussion helps readers gain a clearer understanding of the contribution of this work.

\section{More Implementation Details}\label{app:more details}

For video generation, we train our generator based on the pre-trained street view image generation model from \citet{gao2023magicdrive}.
By adding the proposed relative pose control, we train 4 epochs (77040 steps) on the nuScenes training set with a learning rate of $8e^{-5}$.
We follow the settings for 7-frame videos described in \citet{gao2023magicdrive}, using $224\times 400$ for each view but extending to $T=16$ frames.
Consequently, for 3DGS generation, we select $t=8$ as the canonical space.
Except we change the first 500 steps to optimize $(s_{c,t}, b_{c,t})$ for each view and $\lambda_{\text{reg}_{o}}=1.0$, other settings are the same as 3DGS.

For the monocular depth model, we use ZoeDepth~\citep{bhat2023zoedepth}.
Although it is trained for metric depth estimation, due to domain differences, raw estimation is not usable, as shown in \secref{sec:ablation} and \figref{fig:depth opt}.
Methodologically, \methodname does not rely on a specific depth estimation model.
Better estimations can further improve our scene generation quality.

Since GS only supports perspective rendering, to stitch the view for panorama, we use code provided by \url{https://github.com/timy90022/Perspective-and-Equirectangular} to transform from perspective to equirectangular.

All our experiments are conducted with NVIDIA V100 32GB GPUs.
The generation of a single scene takes about 2 minutes for video generation and 30 minutes for GS generation.
For reference, 3DGS reconstruction typically takes about 23 minutes for scenes of similar scales.
Therefore, the proposed enhancement is efficient.
As for rendering, there is no additional computation for our method compared with 3DGS.

As mentioned in \secref{sec:ablation}, we use two testing scenarios to show the effectiveness of the enhanced GS pipeline as an ablation study, \ie, 360$^\circ$ and vary-t. We use one testing trajectory as an example to illustrate the training/testing view splits in \tabref{tab:settings}.

\input{tables/settings}

\section{Optimization Flow}\label{app:optimflow}
We demonstrate the overall optimization flow of the proposed FTGS in \algref{alg:overall}.
Line 2 is the first optimization of monocular depths.
Lines 4-8 refer to the second optimization of the monocular depths.
Lines 10-16 are the main loop for FTGS, where we consider temporal offsets on Gaussians, camera pose optimization for local inconsistency, and AEs for appearance discrepancies among views.

\input{tables/algorithm}

\input{tables/ablation_gs}

\section{More Ablation Study}\label{app:abl}

\noindent\textbf{Ablation on Offset Choice for Fault-Tolerant GS}.
In addition to the overall module ablation, we observe that for Fault-Tolerant GS, beyond the Gaussians' center coordinates, their attributes (including anisotropic covariance, opacity, and harmonic coefficients) can also be utilized to address local inconsistencies. To verify the effects of different choices, we randomly select 10 scenes from the nuScenes validation set for experimentation. As shown in \tabref{tab:rebuttal}, the results obtained by adding offsets to the center coordinates (xyz) are the best. This aligns with our observation that local inconsistencies in the generated views occur primarily in the shape of objects, and thus, xyz displacements can most effectively resolve these inconsistencies.

\begin{figure}[ht]
    \centering
    \includegraphics[width=0.98\linewidth]{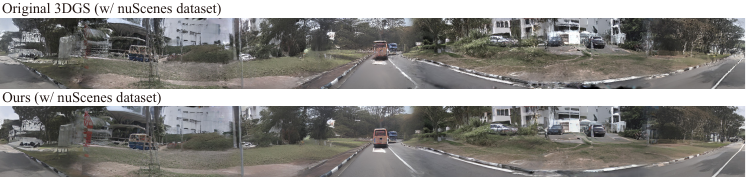}
    \caption{Qualitative Comparison with 3DGS baseline on the same scene shown in \figref{fig:comparison}. With our improved FTGS pipeline, the generated 3DGS contains fewer floaters, leading to higher-quality rendering.}
    \label{fig:comparison-2}
\end{figure}

\noindent\textbf{Qualitative Comparison with 3DGS baseline}.
\figref{fig:comparison-2} shows the results of reconstructing the generated video directly using the 3DGS algorithm. As discussed in \secref{sec:gs}, 3DGS tends to amplify artifacts in the generated view, whereas our improved FTGS pipeline effectively controls the impact of these artifacts, ensuring the quality of the 3D scene generation.

\section{More Reconstruction Baseline}\label{app:4dgs}
Focusing on dynamic scenes, 4DGS~\citep{Wu_2024_CVPR} introduces comprehensive improvements over 3DGS and achieves notable results. Therefore, we replace Fault-Tolerant GS with 4DGS.
As shown in \tabref{tab:4dgs}, by incorporating non-rigid dynamics, 4DGS already performs better than 3DGS.
However, in our task, 4DGS underperforms compared to our Fault-Tolerant GS. Based on the results in \tabref{tab:rebuttal}, we hypothesize that our FTGS algorithm only needs to address local inconsistency caused by content inconsistency. Allocating excessive degrees-of-freedom at the representational level may hinder model convergence, thus 4DGS does not yield better results in our scenarios.

\begin{table}[ht]
\centering
\caption{Comparison with 4DGS~\citep{Wu_2024_CVPR}.
We randomly sample 10 scenes from the nuScenes validation set for experiments and apply color correction (cc) to all the renderings.
}
\label{tab:4dgs}
\begin{tabular}{@{}l|c|c|c|c@{}}
\toprule
Methods & L1 $\downarrow$ & PSNR $\uparrow$ & SSIM $\uparrow$ & LPIPS $\downarrow$ \\ \midrule
3DGS & 0.0733 & 19.7514 & 0.5210 & 0.4496 \\
4DGS & 0.0601 & 21.1195 & 0.5892 & 0.4475 \\ \midrule
\textbf{Ours} & \textbf{0.0546} & \textbf{21.9428} & \textbf{0.6288} & \textbf{0.2759} \\ \bottomrule
\end{tabular}
\end{table}

\section{Limitation and Future Work}
As a data-centric method, \methodname sometimes struggles to generate complex objects like pedestrians, whose appearances are intricate. Additionally, areas with much texture detail (e.g., road fences) or small spatial features (e.g., light poles) are occasionally poorly generated due to limitations in the 3DGS method. 
Future work may focus on addressing these challenges and further improving the quality and robustness of generated 3D scenes.

\section{Comparison with Simple Baselines}\label{app:baseline}
\begin{figure}[ht]
     \centering
     \begin{subfigure}[t]{0.21\textwidth}
         \centering
         \includegraphics[height=1.6cm]{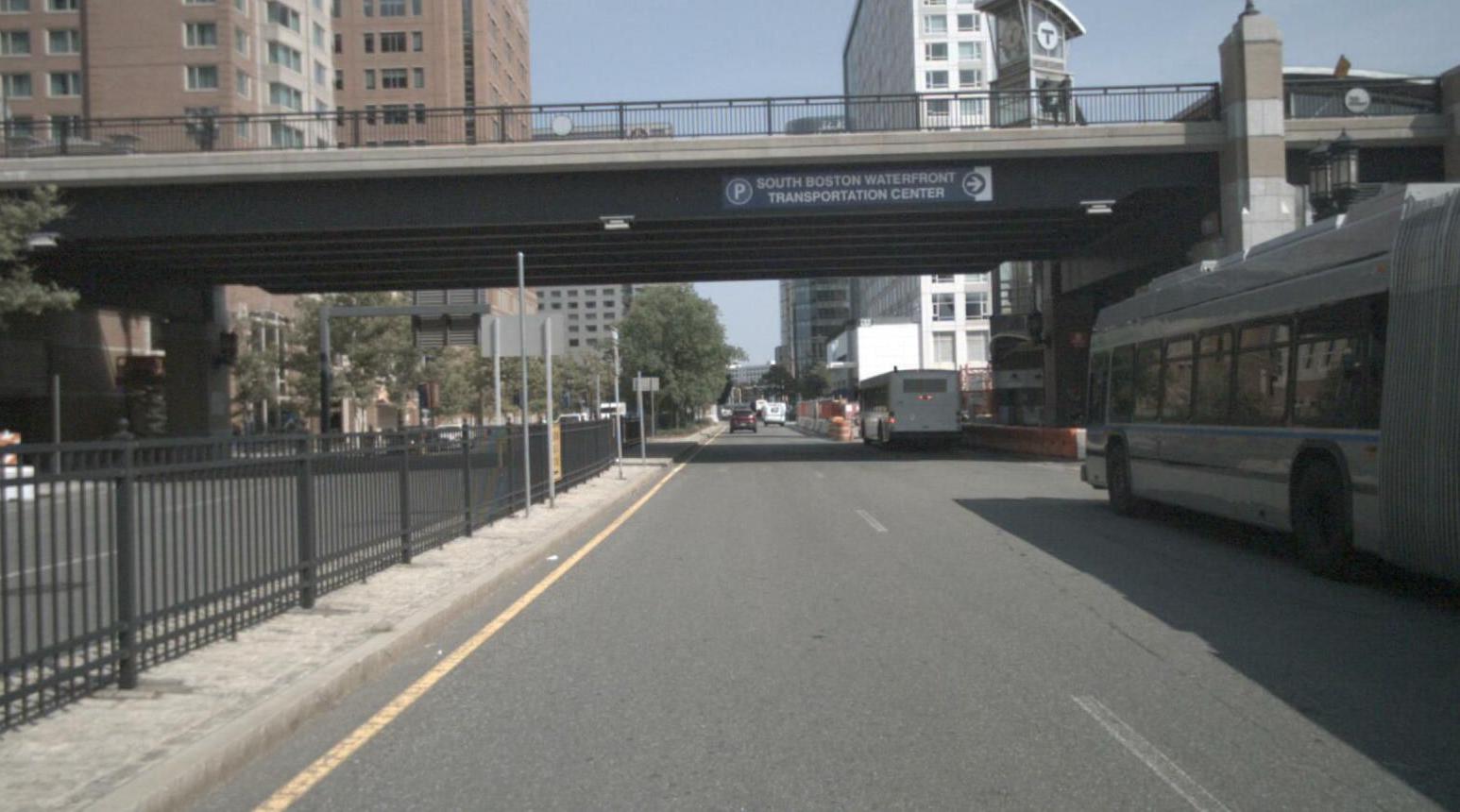}
         \caption{Conditional image to LucidDreamer \citep{chung2023luciddreamer}}
         \label{fig:ld_cond}
     \end{subfigure}
     \hfill
     \begin{subfigure}[t]{0.7\textwidth}
         \centering
         \includegraphics[height=1.6cm]{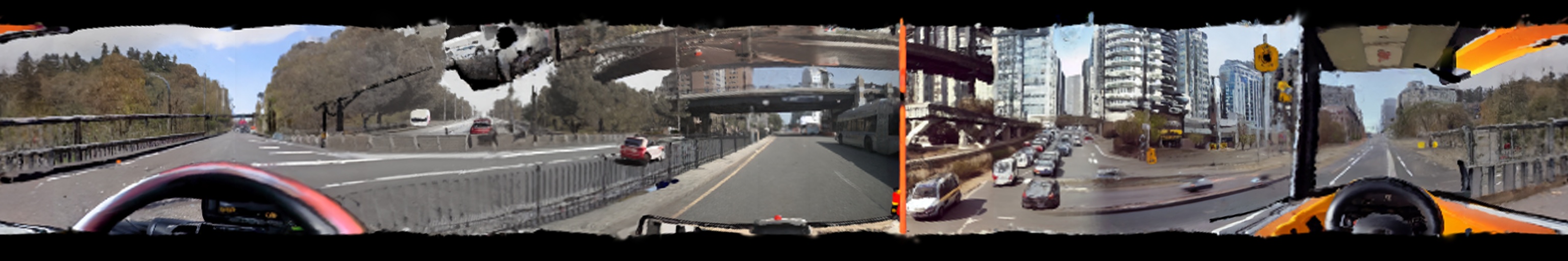}
         \caption{Scene generated by LucidDreamer~\citep{chung2023luciddreamer}, with text ``\textit{A driving scene in the city from the front camera of the vehicle. A bus on the right side. There is a bridge overhead. There is a railing in the center of the left road. Some vehicles ahead}''}
         \label{fig:ld_result}
     \end{subfigure}
     \begin{subfigure}[b]{\textwidth}
         \centering
         \includegraphics[width=\linewidth]{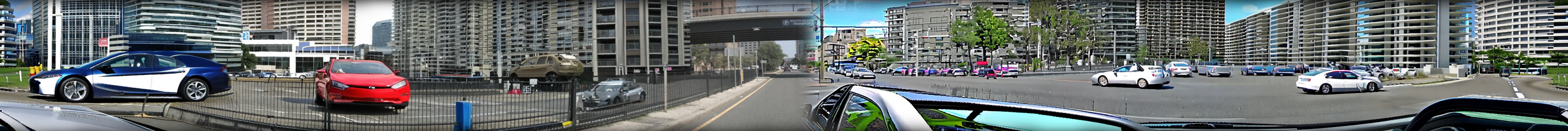}
         \caption{Scene generated by WonderJourney~\citep{yu2024wonderjourney}. We set the focal length to be the same as the conditional image. WonderJourney cannot control road semantics (many objects are physically implausible) and fails in loop closure for 360$^{\circ}$ scene generation. Conditions are the same as \figref{fig:ld_result}.}
         \label{fig:wj_result2}
     \end{subfigure}
     \begin{subfigure}[b]{\textwidth}
         \centering
         \includegraphics[width=\textwidth]{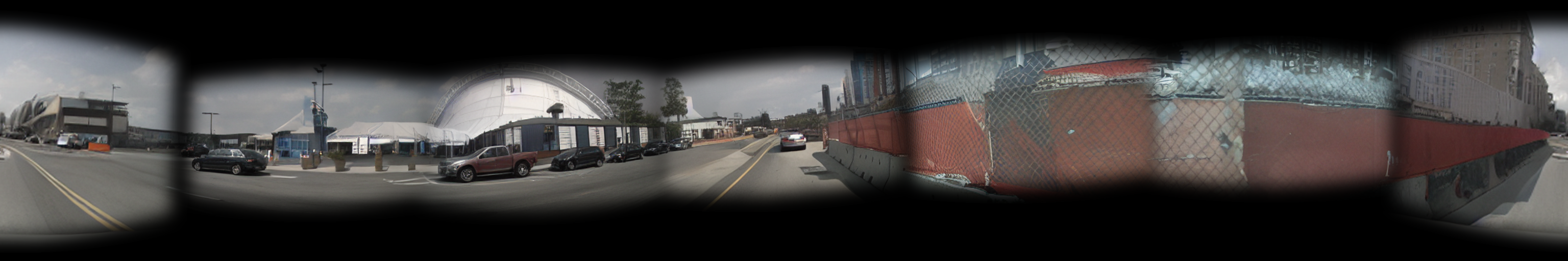}
         \caption{Stitched panorama with real camera views from nuScenes dataset. Due to the limited overlaps, there are many empty (black) areas.}
         \label{fig:real_data}
     \end{subfigure}
     \begin{subfigure}[b]{\textwidth}
         \centering
         \includegraphics[width=\textwidth]{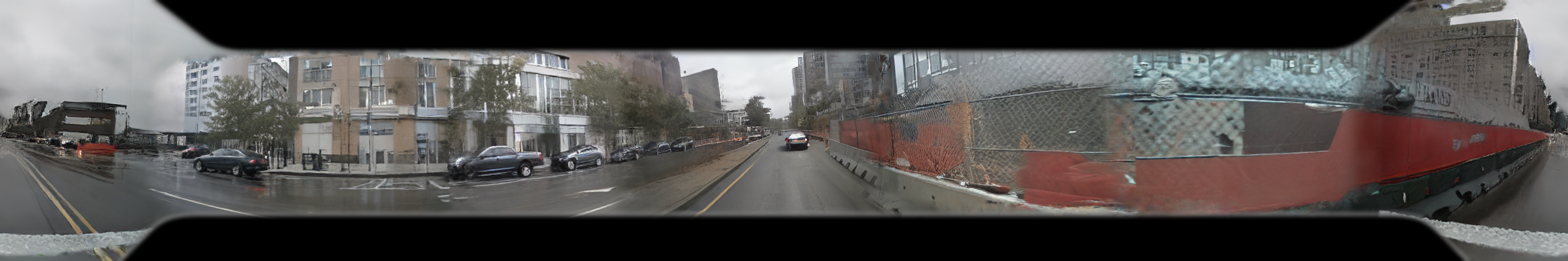}
         \caption{Panorama from \methodname. The scene is generated with the same object boxes and BEV map as \figref{fig:real_data}, but turned to ``\textit{rainy day}''.}
         \label{fig:ours_rain}
     \end{subfigure}
    \caption{Comparison with two baselines (LucidDreamer~\citep{chung2023luciddreamer} and WonderJourney~\citep{yu2024wonderjourney}) and direct stitching real images.}
    \label{fig:simple_baselines}
\end{figure}
As shown in \figref{fig:simple_baselines}, we further compare \methodname with two baselines, \ie, LucidDreamer~\citep{chung2023luciddreamer} and WonderJourney~\citep{yu2024wonderjourney}.
The former method has been proposed recently and takes text description as the only condition.
Thus, it is hard to generate photo-realistic street scenes.
When providing multi-view video frames from nuScenes with known camera poses, their pipeline fails to reconstruct.
We suppose the reason is limited overlaps and errors from depth estimation.
As suggested by the released code, we changed the image generation model to \texttt{lllyasviel/control\_v11p\_sd15\_inpaint} for inpainting by providing a nuScenes image, \ie, \figref{fig:ld_cond}.
However, due to the lack of controllability, the results from LucidDream (\eg, \figref{fig:ld_result}) are unsatisfactory.
On the other hand, due to the lack of control over objects within the scene, WonderJourney struggles to generate coherent scenes. 
Inpainting-based methods like the two above exhibit a pronounced sense of patches and face significant challenges in achieving 360$^{\circ}$ coverage.

\figref{fig:real_data} further shows directly stitching real data.
It is also bad due to the limited overlaps between views.
On the contrary, the scene generated from \methodname can render a continuous panorama, as shown in \figref{fig:ours_rain}, which is also controllable through multiple conditions.

Note that, panorama generation is only one of the applications of our generated scenes.
We show them just for convenient qualitative comparison within the paper.
Since our scene generation contains geometric information, it can be rendered from any camera view, as shown in \figref{fig:comparison}.

\section{Implementation Detail of Appearance Embedding}\label{app:arch}
\begin{figure}[ht]
    \centering
    \includegraphics{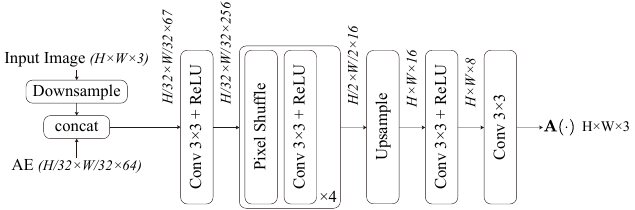}
    \caption{The CNN architecture of appearance modeling, as introduced in \secref{sec:gs}.}
    \label{fig:ae_arch}
\end{figure}

We show in \figref{fig:ae_arch} the detailed architecture of the CNN used in our appearance modeling.
The AE map is $32\times$ smaller than the input image to reduce the computational cost.
Hence, we first downsample the input image by $32\times$.
Then, we use $3\times3$ convolution for feature extraction and pixel shuffle for upsampling.
Each convolution layer is activated by ReLU.

\section{Broader Impacts}\label{app:impact}
The implementation of \methodname in controllable 3D street scene generation could potentially revolutionize the autonomous driving industry.
By creating detailed 3D scenarios, self-driving vehicles can be trained more effectively and efficiently for real-world applications, thereby leading to improved safety and accuracy.
Moreover, it could potentially provide realistic simulations for human-operated vehicle testing and training, thus contributing to reducing the occurrence of accidents on the roads while enhancing driver expertise.
In the broader scope, \methodname could be of considerable value to the virtual reality industry and video gaming industry, enabling these sectors to generate more lifelike 3D scenes and intricate gaming experiences.

On the downside, the development and application of such advanced technology could lead to certain unwanted scenarios. For instance, the increased automation in industries, driven by the potential of this technology, could lead to job losses for drivers and other related professionals as their roles become automated.
A societal transition will be needed to avoid negative impacts on employment levels and the fairness of wealth distribution.

\section{More Qualitative Results}
\begin{figure}[p]
    \centering
    \includegraphics[width=\linewidth]{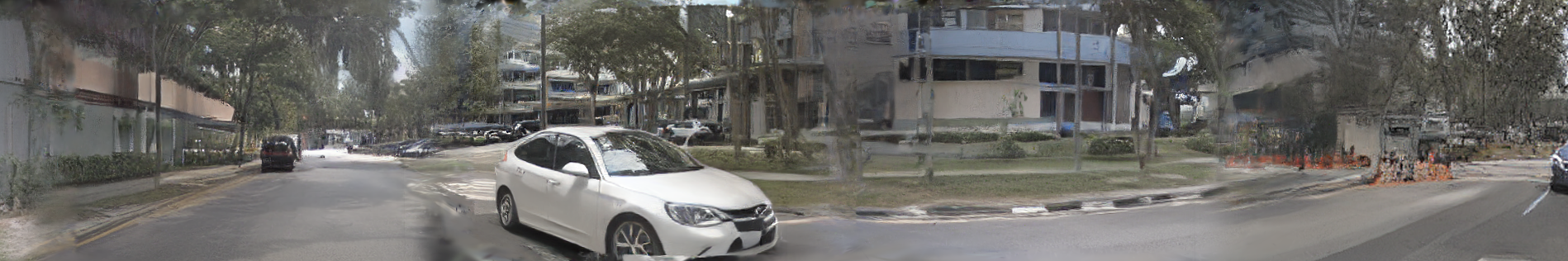}

    \includegraphics[width=\linewidth]{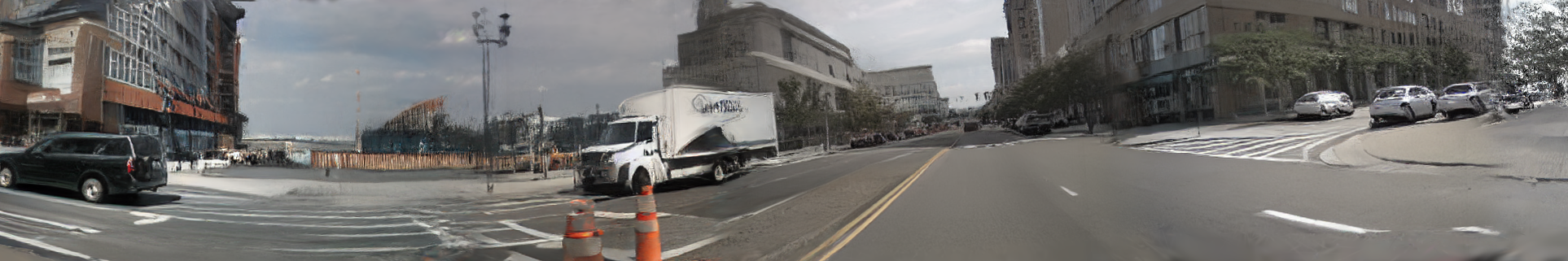}
    
    \caption{Generated street scenes from \methodname.
    We adopt control signals from the nuScenes validation set. We crop the center part for better visualization.}
    \label{fig:more_center}
\end{figure}

\begin{figure}[p]
    \centering
    \includegraphics[width=\linewidth]{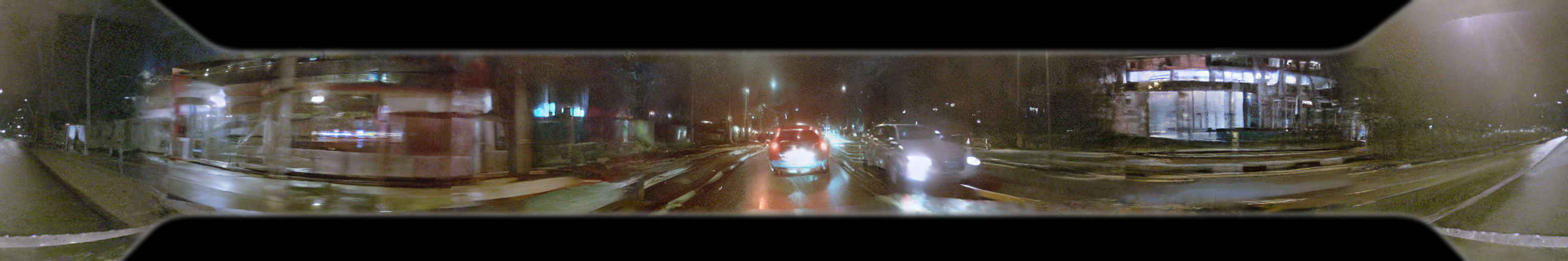}

    \includegraphics[width=\linewidth]{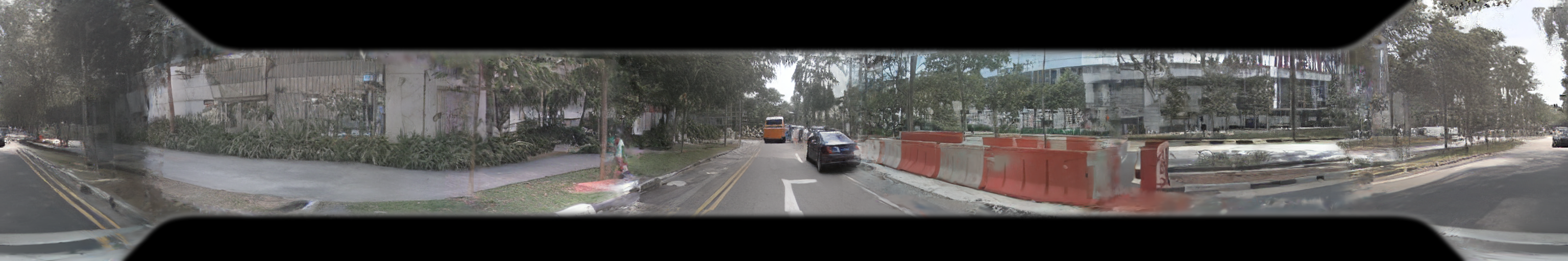}
    \includegraphics[width=\linewidth]{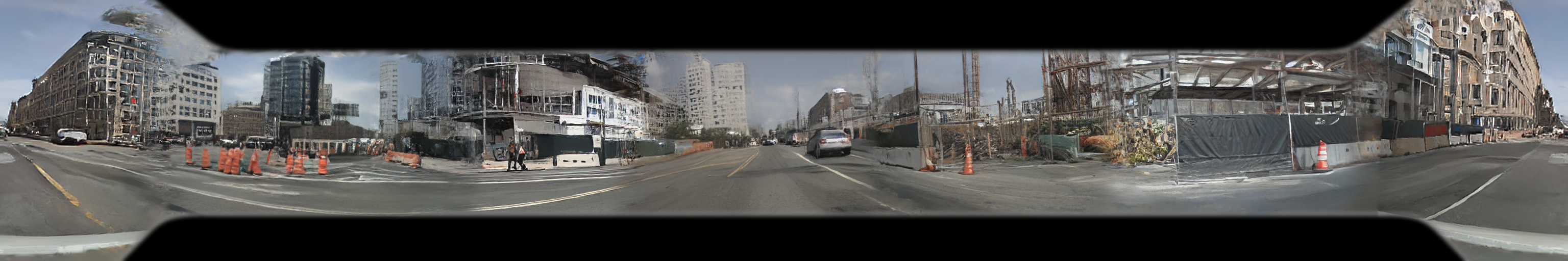}
    
    \includegraphics[width=\linewidth]{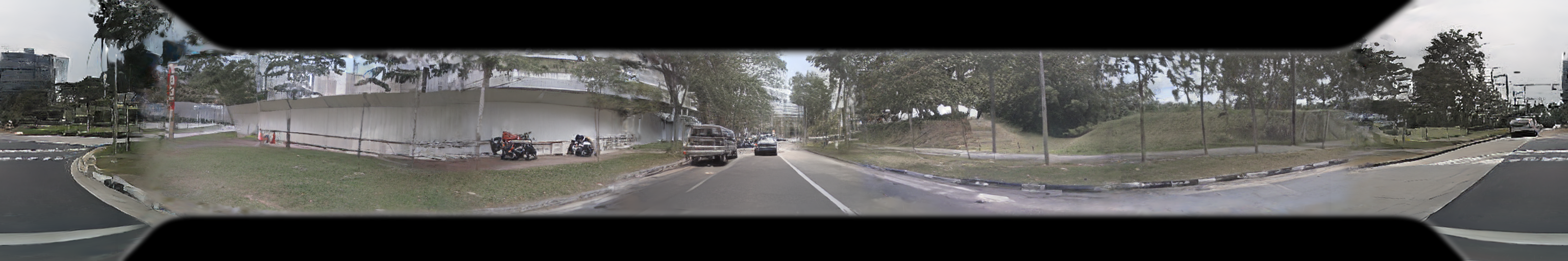}

    \caption{Generated street scenes from \methodname.
    We adopt control signals from nuScenes validation set. The black regions are not fully covered, constrained by the camera's FOV.}
    \label{fig:more}
\end{figure}

We show more generated street scenes from \methodname in \figref{fig:more_center} and \figref{fig:more}.

%% file: tables/settings.tex
\begin{table}[hb]
    \centering
    \setlength\tabcolsep{3.5pt}
    \begin{tabular}{c|c|c|l}
    \hlinew{0.75pt}
    name & \#test & \#train & camera poses  \\
    \hlinew{0.75pt}
    \multirow{2}{*}{360$^\circ$} & \multirow{2}{*}{6} & \multirow{2}{*}{90} & \multirow{2}{*}{\includegraphics[width=5cm]{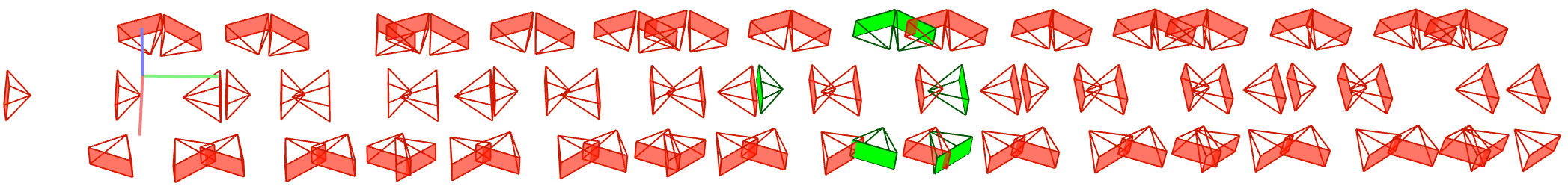}} \\
    &&&\\
    \hline
    \multirow{2}{*}{vary-t} & \multirow{2}{*}{12} & \multirow{2}{*}{84} & \multirow{2}{*}{\includegraphics[width=5cm]{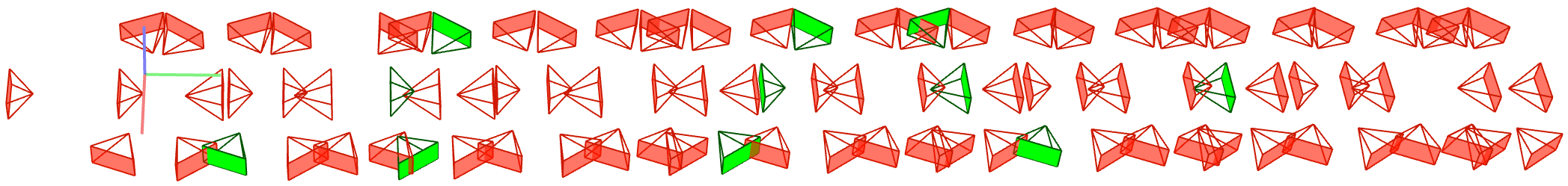}} \\
    &&&\\
    \hlinew{0.75pt}
    \end{tabular}
    \vspace{-0.3cm}
    \caption{
    Two settings for ablation study on the enhanced GS pipeline.
    Testing views are in green while training views are in red.
    }
    \label{tab:settings}
    \vspace{-0.3cm}
\end{table}

%% file: tables/algorithm.tex
\begin{algorithm}[hb]
\caption{Enhanced Fault-Tolerant Gaussian Splatting (FTGS)}
\label{alg:overall}
\begin{algorithmic}[1]
\small
\REQUIRE{camera views $\{\mathcal{I}_{i}\}$, camera parameters $\{\mb{P}^{0}_{i}\}$, monocular depth $\{\mathcal{D}_{i}\}$, optimization steps for depth $s_{D}$, camera pose $s_{\text{C}}$, and GS $s_{\text{GS}}$}
\ENSURE{FTGS of the scene $\{\pmb{\mu}_{p},\pmb{\mu}^{o}_{p},\pmb{\Sigma}_{p},\text{SH}_{p}\}$, and optimized camera pose $\{\mb{P}^{0}_{i}\}$}

\STATE  $\mathcal{P}_{SfM}=$ PCD from SfM
\STATE  Optimize $(s_{c,t}, b_{c,t})$ with $\mathcal{P}_{\text{SfM}}$ for each $\{c,t\}$
\STATE Random initialize AEs $\{\mb{e}_{i}\}$
\FOR{step in $1, \dots, s_{D}$}
    \STATE Random pick one view $\mathcal{I}_{i}$
    \STATE $\mathcal{L}=\mathcal{L}_{\text{AEGS}}(\mathcal{I}_{i},\mathcal{I}^{r}_{i},\mb{e}_{i})$
    \STATE Update $(s, b),\mb{e}_{i},\pmb{\Sigma},\text{SH}$ with $\nabla\mathcal{L}$
\ENDFOR
\STATE Initialize $\pmb{\mu}$ with $(s, b)$ and $\mathcal{D}$
\FOR{step in $s_{D}, \dots, s_{\text{GS}}$}
    \STATE Random pick one view $\mathcal{I}_{i}$ and get its $t$
    \STATE $\mathcal{L}=\mathcal{L}_{\text{FTGS}}(\mathcal{I}_{i},\mathcal{I}^{r}_{i},\mb{e}_{i},\pmb{\mu}^{o}(t))$
    \STATE Update $\pmb{\mu},\pmb{\mu}^{o}(t),\mb{e}_{i},\pmb{\Sigma},\text{SH}$ with $\nabla\mathcal{L}$
    \IF{step $>s_{\text{C}}$}
        \STATE Update $\mb{P}^{0}_{i}$ with $\nabla\mathcal{L}$
    \ENDIF
\ENDFOR
\end{algorithmic}
\end{algorithm}

%% file: tables/ablation_gs.tex
\begin{table}[ht]
\setlength\tabcolsep{3.5pt}
\centering
\caption{Ablating comparison with offsets on anisotropic covariance (Cov.), opacity, and harmonic coefficients (Features) properties in GS.
We randomly sample 10 scenes from the nuScenes validation set for experiments and apply color correction (cc) to all the renderings.
}
\label{tab:rebuttal}
\begin{tabular}{@{}l|c|c|c|c@{}}
\toprule
Methods & L1 $\downarrow$ & PSNR $\uparrow$ & SSIM $\uparrow$ & LPIPS $\downarrow$ \\ \midrule
3DGS & 0.0733 & 19.7514 & 0.5210 & 0.4496 \\ \midrule
Features offset & 0.0624 & 20.9882 & 0.5940 & 0.3463 \\
Cov. offset & 0.0632 & 20.8133 & 0.5854 & 0.3626 \\
Opacity offset & 0.0656 & 20.5332 & 0.5733 & 0.3845 \\ \midrule
\textbf{Ours (xyz offset)} & \textbf{0.0546} & \textbf{21.9428} & \textbf{0.6288} & \textbf{0.2759} \\ \bottomrule
\end{tabular}
\end{table}